\documentclass{article}
\PassOptionsToPackage{numbers, compress}{natbib}

\usepackage[final]{neurips_2022}    

\usepackage[utf8]{inputenc} 
\usepackage[T1]{fontenc}    
\usepackage{hyperref}       
\usepackage{url}            
\usepackage{booktabs}       
\usepackage{amsfonts}       
\usepackage{nicefrac}       
\usepackage{microtype}      
\usepackage{xcolor}         
\usepackage{tikz}
\usepackage{caption}
\usepackage{subcaption}
\usepackage{float}

\title{Can Strategic Data Collection Improve\\
the Performance of Poverty Prediction Models?}

\author{%
Satej Soman\\
School of Information \& Global Policy Lab\\
University of California, Berkeley \\
Berkeley, CA 94704 \\
\texttt{satej@berkeley.edu} \\
\And
Emily Aiken \\
School of Information \& Global Policy Lab\\
University of California, Berkeley \\
Berkeley, CA 94704 \\
\texttt{emilyaiken@berkeley.edu} \\
\AND
Esther Rolf \\
Harvard Data Science Initiative \& \\  Center for Research on Computation and Society \\
Harvard University \\
Cambridge, MA \\
\texttt{erolf@seas.harvard.edu} \\
\And
Joshua Blumenstock \\
School of Information, Goldman School of Public Policy\\\& Global Policy Lab \\
University of California, Berkeley \\
Berkeley, CA 94704 \\
\texttt{jblumenstock@berkeley.edu} \\
}

\begin{document}
\maketitle

\begin{abstract}
Machine learning-based estimates of poverty and wealth are increasingly being used to guide the targeting of humanitarian aid and the allocation of social assistance.
However, the ground truth labels used to train these models are  typically borrowed from existing surveys that were designed to produce national statistics -- not to train machine learning models.
Here, we test whether adaptive sampling strategies for ground truth data collection can improve the performance of poverty prediction models.
Through simulations, we compare the status quo sampling strategies (uniform at random and stratified random sampling) 
to alternatives that prioritize acquiring training data based on model uncertainty or model performance on sub-populations.
Perhaps surprisingly, we find that none of these active learning methods improve over uniform-at-random sampling. 
We discuss how these results can help shape future efforts to refine machine learning-based estimates of poverty.

\end{abstract}

\section{Introduction}
Over the past decade, several influential studies have demonstrated how machine learning (ML) -- when applied to digital data sources -- can provide accurate, timely, and granular estimates of wealth and poverty \cite{blumenstock2015predicting,jean2016combining,steele_mapping_2017,blumenstock2018estimating,burke_using_2020,yeh2020using,chi_microestimates_2022}. 
These ML-based estimates, which can be produced at a fraction of the cost of traditional survey-based methods of data collection, can fill in data gaps in poor, fragile, and conflict-affected regions where recent surveys do not exist \cite{jerven2013poor}. And increasingly, the ML-based estimates are being used to inform critical policy decisions, including the targeting of humanitarian aid \cite{aiken2022machine} and the allocation of social assistance \cite{smythe_geographic_2022,gentilini2021cash}.


Despite the important role ML-based poverty estimates now play in development policy and disaster response, very little consideration has been given to the sampling strategy used to collect the ground truth labels used to train the ML models. Instead, the vast majority of poverty prediction models are trained on  publicly-available household surveys that were collected for a different purpose (typically, to provide nationally representative demographic and health statistics). 
These standardized surveys typically use (stratified) random sampling to 
reduce measurement error \citep{howes1998does}. However, sampling training data according to these strategies may not optimize predictive power for a machine learning algorithm subject to a budget constraint for data collection. In an \textit{adaptive sampling} strategy, each incremental round of data acquisition is informed by an aspect of the performance of the model trained on data seen in previous rounds of data collection. In this paper, we investigate whether adaptive sampling strategies for label data collection improve the performance of machine learning models trained to predict poverty from digital data sources, focusing on model accuracy and fairness.

\section{Application Context}
This work builds on an emergency cash-transfer program run by the government of Togo that used poverty estimates from digital data sources to prioritize the distribution of emergency mobile money payments in response to the COVID-19 pandemic (described in more detail in Appendix \ref{additional_background} and \citet{aiken2022machine}). In designing the program, the government sought to distribute cash to 60,000 of the country's poorest individuals. At the time, however, the government did not have any traditional poverty data with which  to determine eligibility. Instead, the Togolese government turned to satellite imagery analysis to identify the country's poorest districts, and then used machine learning models using features derived from subscribers' mobile phone records to identify the poorest subscribers in eligible districts for humanitarian aid. The ground truth labels for the wealth prediction models came from surveys conducted in Togo prior to the program's launch. An evaluation of the program's targeting found that relative to alternative poverty targeting methods (such as geographic or occupation-based targeting), the phone-based approach was most effective at targeting cash transfers to the poorest people in the country \cite{aiken2022machine}. 

In this paper, we study whether using an adaptive strategy for sampling the poverty labels obtained from the phone survey in Togo could have improved the accuracy of the trained machine learning model. Our results speak most directly to recent efforts to estimate wealth and poverty from mobile phone data. More broadly, these results can inform the larger community of researchers and policymakers interested in estimating social and demographic characteristics from digital data in contexts where traditional data are unavailable or out-of-date. 

\section{Methods}
 We conduct experiments to simulate various data acquisition strategies using an in-person household survey conducted in Togo in 2018 ($N$ = 4,595) with a stratified random sampling strategy \footnote{While the survey was stratified by region and urban/rural designation, our experiment treats this survey as if it were sampled uniform-at-random.}. The survey collected information on daily per-capita consumption for each household interviewed (measured in USD/day), which we use as our ground truth measure of poverty.  We match the household survey to anonymized features derived from call data records provided by the nation’s two cellular network operators. After partitioning the survey-labelled features into a label pool ($N_{\small{\textrm{LP}}}$ = 0.75$N$ = 3,446) and a holdout validation set ($N_{\small{\textrm{V}}}$ = 0.25$N$ = 1,149), we train random forest regressor models \cite{scikit-learn} on subsets of the label pool of incrementally increasing size to simulate how the resulting model performs throughout the data acquisition process. The model's performance is evaluated against the holdout set across all dataset sizes.

The baseline of acquiring data by uniform random sampling is compared against two adaptive data acquisition approaches, in which each data point still remaining in the label pool is assigned a heuristic sampling weight. In the \textit{query-by-committee} strategy, this weight is proportional to the variance of the outputs of the individual decision trees within the random forest regressor; unlabeled points for which the ensemble's constituent models disagree the most are upweighted in the next round of data acquisition. An alternative is \textit{margin-based uncertainty sampling}, in which an independent logistic regression classifier is trained on the same features to predict whether a given individual falls below a relevant poverty threshold; points whose probability of classification is closer to 50\% are upweighted in order to gain more information on points whose classification is close to random. Query-by-committee and uncertainty sampling are two standard methods for active learning for model accuracy \citep{settles2009active}. They have both been shown to substantially improve model performance subject to a budget constraint for data collection in small and relatively low-dimensional datasets unrelated to the poverty prediction task we consider in this paper \citep{settles2009active, settles2008analysis, mccallumzy1998employing, korner2006multi}. 

Inspired by literature on active learning for fairness outcomes, \cite{abernethy2020active, cai2022adaptive}, we also consider adaptively constructing the training dataset by increasing the sampling weights for the points from the demographic categories with the lowest accuracy (\textit{accuracy-weighted}),  highest mean-squared-error (\textit{MSE-weighted}), or highest exclusion/inclusion error rates (\textit{disparity-weighted}). In these group-based strategies, each member of the demographic category is assigned the same weight, but each group's weight differs by group-specific model performance in each round of simulated data collection.

Focusing on these demographic category weights also allows us to assess aspects of fairness of the model during the data collection process, as measured by minimum group accuracy, maximum group loss, or total inclusion-exclusion errors. These goals are independent of increasing the model's predictive power: even if deviation from uniform-at-random sampling provides no benefits in increasing accuracy or reducing MSE, there may be benefits in model fairness. In this investigation, the demographic category is geographic residence in each of the major admin-1 regions of Togo (depicted in more detail in Figure \ref{fig:togo_map}), inspired by previous analysis indicating there may exist slight geographic bias to the phone-based models \cite{aiken2022machine}.

We run 50 simulations for each strategy, with incremental dataset sizes $S_t$ logarithmically spaced between 0 and the full label pool size to better illustrate the effects of different strategies at smaller sample sizes. Within each simulation, the newly sampled points are appended to the previously-seen dataset ($S_t - S_{t-1}$ new data points are seen at time $t$ rather than $S_t$ points sampled memorylessly). Our code for the experiments is available on \href{https://github.com/satejsoman/cdr-active-learning}{GitHub}.

\section{Results}
\begin{figure}[th]
\centering
\includegraphics[width=0.75\textwidth]{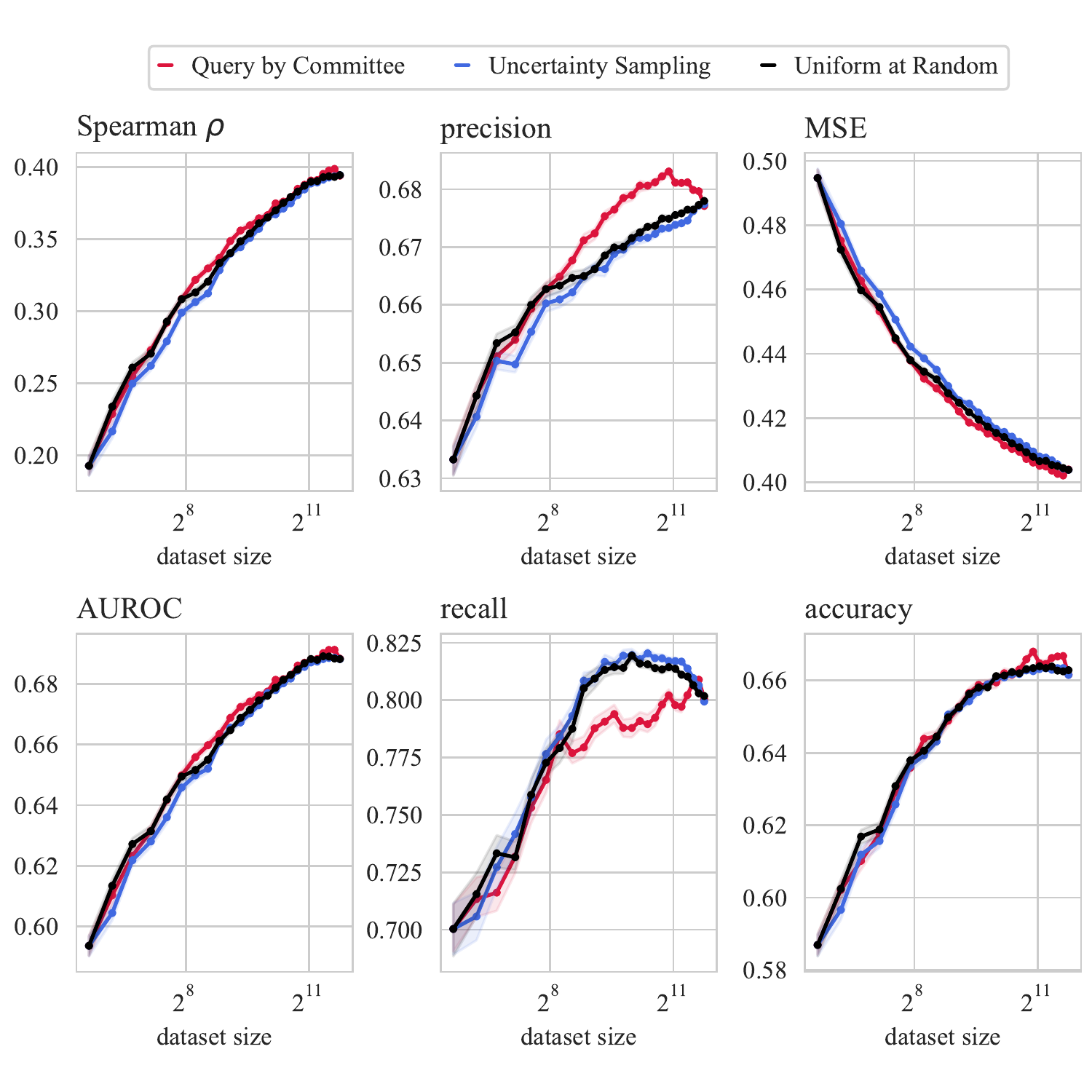}
\caption{Comparison of model-based data acquisition strategies on key machine learning performance metrics as dataset size incrementally grows.}
\label{fig:point_sampling_metrics}
\end{figure}

 We present the results of each of our sampling strategies, comparing individual level active learning methods (query-by-committee and uncertainty sampling) and group-level active learning methods (accuracy, MSE, and disparity weighting) to a baseline of uniform-at-random sampling. 
 In Figure \ref{fig:point_sampling_metrics}, we report the average performance of each sampling method across 50 simulations, with shaded regions representing boostrapped 95\% confidence intervals. Evaluating across a number of standard machine learning accuracy metrics (Spearman $\rho$, mean-square-error, AUROC, accuracy, precision, and recall).\footnote{Poverty predictions are continuous and produced with regression models; we binarize predictions using the international poverty line of US\$1.90/day to obtain binary classification metrics.}, it is clear that neither model-based nor group-weighting strategies provide substantial gains to overall model performance above uniform at random sampling. With some strategies, such as the query by committee approach, we do see some gains to model precision, but these are on the order of 1\% and are offset by reduction in recall (which is a more important metric for aid prioritization, since a model with lower recall will mistakenly exclude more recipients who should receive aid).

\begin{figure}[ht]
\centering
\includegraphics[width=\textwidth]{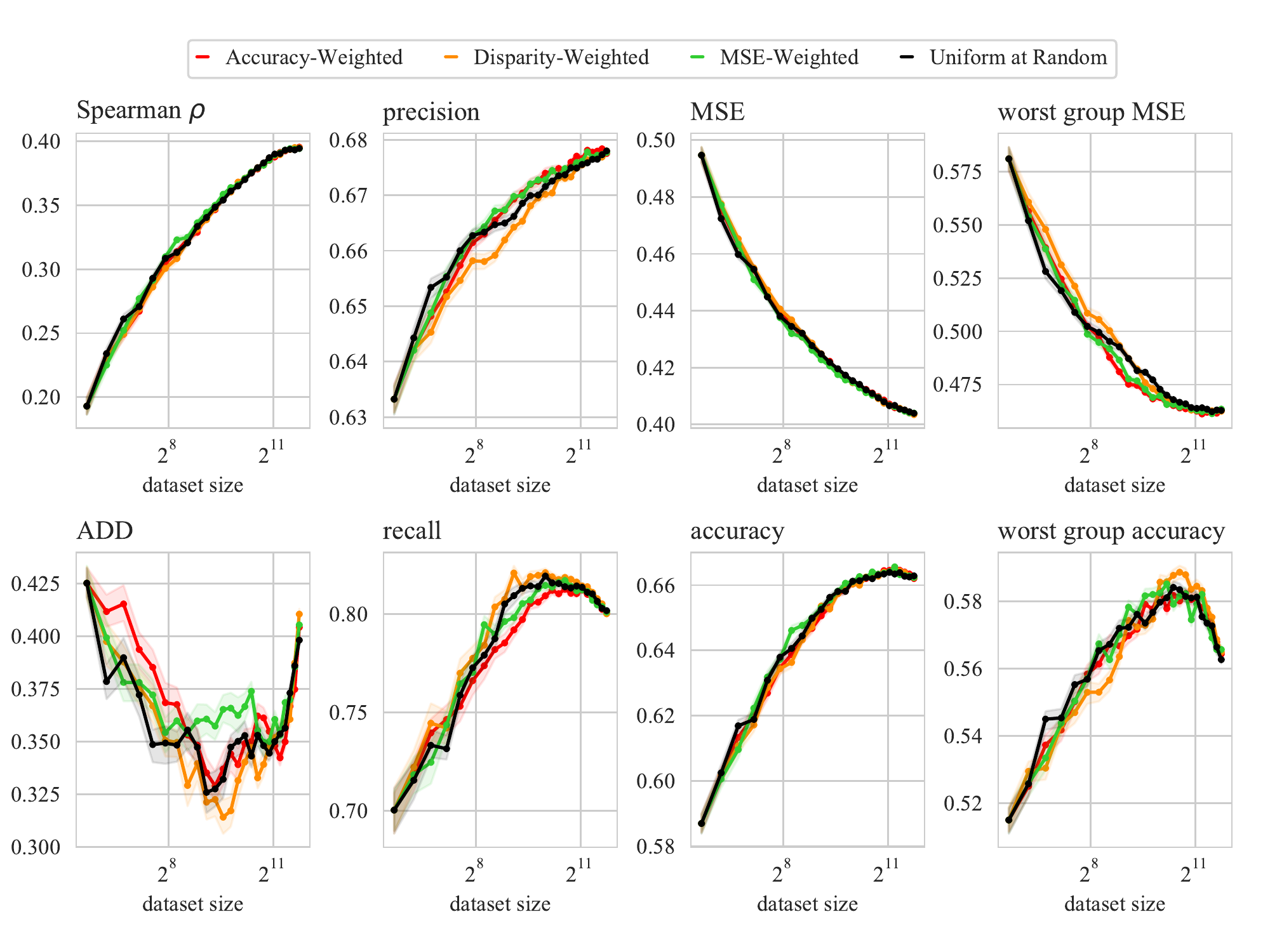}
\caption{Comparison of group weighting strategies on standard machine learning metrics, as well as fairness-oriented metrics such as minimum group-level accuracy, maximum group-level mean-square-error, and total inclusion/exclusion errors (ADD).}
\label{fig:group_sampling_metrics}
\end{figure}

Our group-based sampling strategies rely on upweighting groups (regions of Togo) for which our model performs worse. We evaluate these strategies based on the same metrics as above, and three additional fairness-related metrics calculated for the major geographic regions in Togo. In Figure \ref{fig:group_sampling_metrics}, we plot minimum group accuracy, maximum group loss, and a summary statistic for targeting inclusion/exclusion errors (absolute demographic deviation; ADD, explained in more detail in Appendix \ref{fairness_metrics_explained}). Mean values are plotted, with shaded regions representing bootstrapped 95\% confidence intervals. Again, we observe negligible differences in fairness between our active learning approaches and the uniform-at-random baseline sampling strategy. 

Additional experiments in Appendix \ref{app:additional_experiments} detail similar results when using a feature set of reduced dimensionality and a hyperparameter tuning procedure across dataset sizes.

\section{Discussion}
Our results indicate that a well-implemented, uniformly-sampled  survey may be close to optimal as a source of training data for machine learning models that aim to predict poverty levels from digital trace data. In particular, neither the model-based data acquisition strategies nor the group characteristic weighting approaches we considered showed appreciable gains on accuracy or fairness metrics over simple uniform random sampling.

This result has important implications for the growing research literature -- and increasing number of policy applications -- that rely on estimates of poverty predicted from digital trace data. Nearly all such estimates have been generated from ``found'' data, i.e., survey data that were collected for another purpose, and which were subsequently re-used as labels for training a poverty prediction algorithm \cite[e.g.,][]{blumenstock2015predicting,jean2016combining,steele_mapping_2017,blumenstock2018estimating,burke_using_2020,yeh2020using,chi_microestimates_2022}. Our hypothesis in initiating this  analysis was that such re-use was sub-optimal; if the poverty prediction use case was known \textit{ex ante}, we expected there to be performance gains from a more strategic approach to obtaining training data. However, our results indicate that --- at least for the adaptive approaches that we tested -- no such gains exist.

Whether this is a general finding, or one that is specific to our particular data context (i.e., mobile phone data from Togo), is an area of future work that we are continuing to explore. We also see several new directions in which to extend this analysis. For instance, all of the simulations described here assume an identical labeling cost for all data points. This uniform-cost model is appropriate in settings where the cost of interviewing an individual is the same regardless of their physical location in the country -- as is the case with phone-based surveys. By contrast, many settings involve highly heterogeneous sampling costs -- such as household surveys where some households are easily accessed and others are remote. In such settings, factoring in the costs of traveling to survey locations and conducting surveys may provide more realistic strategies for collecting training labels in household surveys. In the current investigation, in which groups are defined geographically, group-specific sampling is especially cost-infeasible because where the group to be sampled may change in every round of incremental data acquisition. In continued work, we plan to investigate cost-sensitive data acquisition strategies, and generate theoretical explanations for the lack of gains in any of the non-uniform sampling strategies.


\newpage
\begin{ack}
This material is based upon work supported by the National Science Foundation under Grant No. DGE- 2125913.
\end{ack}
\bibliographystyle{unsrtnat}
\bibliography{main}

\begin{thebibliography}{20}
\providecommand{\natexlab}[1]{#1}
\providecommand{\url}[1]{\texttt{#1}}
\expandafter\ifx\csname urlstyle\endcsname\relax
  \providecommand{\doi}[1]{doi: #1}\else
  \providecommand{\doi}{doi: \begingroup \urlstyle{rm}\Url}\fi

\bibitem[Blumenstock et~al.(2015)Blumenstock, Cadamuro, and
  On]{blumenstock2015predicting}
Joshua~E Blumenstock, Gabriel Cadamuro, and Robert On.
\newblock Predicting poverty and wealth from mobile phone metadata.
\newblock \emph{Science}, 350\penalty0 (6264):\penalty0 1073--1076, 2015.

\bibitem[Jean et~al.(2016)Jean, Burke, Xie, Davis, Lobell, and
  Ermon]{jean2016combining}
Neal Jean, Marshall Burke, Michael Xie, W~Matthew Davis, David~B Lobell, and
  Stefano Ermon.
\newblock Combining satellite imagery and machine learning to predict poverty.
\newblock \emph{Science}, 353\penalty0 (6301):\penalty0 790--794, 2016.

\bibitem[Steele et~al.(2017)Steele, Sundsøy, Pezzulo, Alegana, Bird,
  Blumenstock, Bjelland, Engø-Monsen, Montjoye, Iqbal, Hadiuzzaman, Lu,
  Wetter, Tatem, and Bengtsson]{steele_mapping_2017}
Jessica~E. Steele, Pål~Roe Sundsøy, Carla Pezzulo, Victor~A. Alegana,
  Tomas~J. Bird, Joshua~Evan Blumenstock, Johannes Bjelland, Kenth
  Engø-Monsen, Yves-Alexandre~de Montjoye, Asif~M. Iqbal, Khandakar~N.
  Hadiuzzaman, Xin Lu, Erik Wetter, Andrew~J. Tatem, and Linus Bengtsson.
\newblock Mapping poverty using mobile phone and satellite data.
\newblock \emph{Journal of The Royal Society Interface}, 14\penalty0
  (127):\penalty0 20160690, February 2017.
\newblock ISSN 1742-5689, 1742-5662.
\newblock \doi{10.1098/rsif.2016.0690}.
\newblock URL
  \url{http://rsif.royalsocietypublishing.org/content/14/127/20160690}.

\bibitem[Blumenstock(2018)]{blumenstock2018estimating}
Joshua~E Blumenstock.
\newblock Estimating economic characteristics with phone data.
\newblock In \emph{AEA papers and proceedings}, volume 108, pages 72--76, 2018.

\bibitem[Burke et~al.(2020)Burke, Driscoll, Lobell, and
  Ermon]{burke_using_2020}
Marshall Burke, Anne Driscoll, David Lobell, and Stefano Ermon.
\newblock Using {Satellite} {Imagery} to {Understand} and {Promote}
  {Sustainable} {Development}.
\newblock Technical Report w27879, National Bureau of Economic Research,
  October 2020.
\newblock URL \url{https://www.nber.org/papers/w27879}.

\bibitem[Yeh et~al.(2020)Yeh, Perez, Driscoll, Azzari, Tang, Lobell, Ermon, and
  Burke]{yeh2020using}
Christopher Yeh, Anthony Perez, Anne Driscoll, George Azzari, Zhongyi Tang,
  David Lobell, Stefano Ermon, and Marshall Burke.
\newblock Using publicly available satellite imagery and deep learning to
  understand economic well-being in {Africa}.
\newblock \emph{Nature communications}, 11\penalty0 (1):\penalty0 1--11, 2020.

\bibitem[Chi et~al.(2022)Chi, Fang, Chatterjee, and
  Blumenstock]{chi_microestimates_2022}
Guanghua Chi, Han Fang, Sourav Chatterjee, and Joshua~E. Blumenstock.
\newblock Microestimates of wealth for all low- and middle-income countries.
\newblock \emph{Proceedings of the National Academy of Sciences}, 119\penalty0
  (3), January 2022.
\newblock ISSN 0027-8424, 1091-6490.
\newblock \doi{10.1073/pnas.2113658119}.
\newblock URL \url{https://www.pnas.org/content/119/3/e2113658119}.
\newblock ISBN: 9782113658118 Publisher: National Academy of Sciences Section:
  Social Sciences.

\bibitem[Jerven(2013)]{jerven2013poor}
Morten Jerven.
\newblock Poor numbers.
\newblock In \emph{Poor Numbers}. Cornell University Press, 2013.

\bibitem[Aiken et~al.(2022)Aiken, Bellue, Karlan, Udry, and
  Blumenstock]{aiken2022machine}
Emily Aiken, Suzanne Bellue, Dean Karlan, Chris Udry, and Joshua~E Blumenstock.
\newblock Machine learning and phone data can improve targeting of humanitarian
  aid.
\newblock \emph{Nature}, 603\penalty0 (7903):\penalty0 864--870, 2022.

\bibitem[Smythe and Blumenstock(2022)]{smythe_geographic_2022}
Isabella~S. Smythe and Joshua~E. Blumenstock.
\newblock Geographic microtargeting of social assistance with high-resolution
  poverty maps.
\newblock \emph{Proceedings of the National Academy of Sciences}, 119\penalty0
  (32):\penalty0 e2120025119, August 2022.
\newblock \doi{10.1073/pnas.2120025119}.
\newblock URL \url{https://www.pnas.org/doi/10.1073/pnas.2120025119}.
\newblock Publisher: Proceedings of the National Academy of Sciences.

\bibitem[Gentilini et~al.(2021)Gentilini, Khosla, and
  Almenfi]{gentilini2021cash}
Ugo Gentilini, Saksham Khosla, and Mohamed Almenfi.
\newblock Cash in the city.
\newblock Technical report, World Bank, 2021.

\bibitem[Howes and Lanjouw(1998)]{howes1998does}
Stephen Howes and Jean~Olson Lanjouw.
\newblock Does sample design matter for poverty rate comparisons?
\newblock \emph{Review of Income and Wealth}, 44\penalty0 (1):\penalty0
  99--109, 1998.

\bibitem[Pedregosa et~al.(2011)Pedregosa, Varoquaux, Gramfort, Michel, Thirion,
  Grisel, Blondel, Prettenhofer, Weiss, Dubourg, Vanderplas, Passos,
  Cournapeau, Brucher, Perrot, and Duchesnay]{scikit-learn}
F.~Pedregosa, G.~Varoquaux, A.~Gramfort, V.~Michel, B.~Thirion, O.~Grisel,
  M.~Blondel, P.~Prettenhofer, R.~Weiss, V.~Dubourg, J.~Vanderplas, A.~Passos,
  D.~Cournapeau, M.~Brucher, M.~Perrot, and E.~Duchesnay.
\newblock Scikit-learn: Machine learning in {P}ython.
\newblock \emph{Journal of Machine Learning Research}, 12:\penalty0 2825--2830,
  2011.

\bibitem[Settles(2009)]{settles2009active}
Burr Settles.
\newblock Active learning literature survey.
\newblock Technical report, University of Wisconsin-Madison, 2009.

\bibitem[Settles and Craven(2008)]{settles2008analysis}
Burr Settles and Mark Craven.
\newblock An analysis of active learning strategies for sequence labeling
  tasks.
\newblock In \emph{proceedings of the 2008 conference on empirical methods in
  natural language processing}, pages 1070--1079, 2008.

\bibitem[McCallumzy and Nigamy(1998)]{mccallumzy1998employing}
Andrew~Kachites McCallumzy and Kamal Nigamy.
\newblock Employing em and pool-based active learning for text classification.
\newblock In \emph{Proc. International Conference on Machine Learning (ICML)},
  pages 359--367. Citeseer, 1998.

\bibitem[K{\"o}rner and Wrobel(2006)]{korner2006multi}
Christine K{\"o}rner and Stefan Wrobel.
\newblock Multi-class ensemble-based active learning.
\newblock In \emph{European conference on machine learning}, pages 687--694.
  Springer, 2006.

\bibitem[Abernethy et~al.(2020)Abernethy, Awasthi, Kleindessner, Morgenstern,
  Russell, and Zhang]{abernethy2020active}
Jacob Abernethy, Pranjal Awasthi, Matth{\"a}us Kleindessner, Jamie Morgenstern,
  Chris Russell, and Jie Zhang.
\newblock Active sampling for min-max fairness.
\newblock \emph{arXiv preprint arXiv:2006.06879}, 2020.

\bibitem[Cai et~al.(2022)Cai, Encarnacion, Chern, Corbett-Davies, Bogen,
  Bergman, and Goel]{cai2022adaptive}
William Cai, Ro~Encarnacion, Bobbie Chern, Sam Corbett-Davies, Miranda Bogen,
  Stevie Bergman, and Sharad Goel.
\newblock Adaptive sampling strategies to construct equitable training
  datasets.
\newblock \emph{arXiv preprint arXiv:2202.01327}, 2022.

\bibitem[\textcommabelow{T}ifrea et~al.(2022)\textcommabelow{T}ifrea, Clarysse,
  and Yang]{tifreauniform}
Alexandru \textcommabelow{T}ifrea, Jacob Clarysse, and Fanny Yang.
\newblock Uniform versus uncertainty sampling: When being active is less
  efficient than staying passive.
\newblock In \emph{Workshop on Adaptive Experimental Design and Active Learning
  in the Real World}, 2022.

\end{thebibliography}

\appendix
\renewcommand{\thefigure}{A\arabic{figure}}
\setcounter{figure}{0}

\newpage
\section*{\huge{Supplementary Materials}}

\section{Additional Background}\label{additional_background}
Togo is a West African country with a population of 8 million, over half of whom live below with international poverty line. In March 2020, the government imposed a lockdown order within the country in response to the first confirmed COVID-19 cases in the country. These lockdowns proved economically disruptive, with risks of food insecurity rising as work became scarce for Togolese citizens employed informally \cite{aiken2022machine}. 

\begin{figure}[ht]
\centering
\begin{subfigure}[b]{0.39\textwidth}
\includegraphics[width=\textwidth,trim=0 -3cm 0 1.35cm,clip]{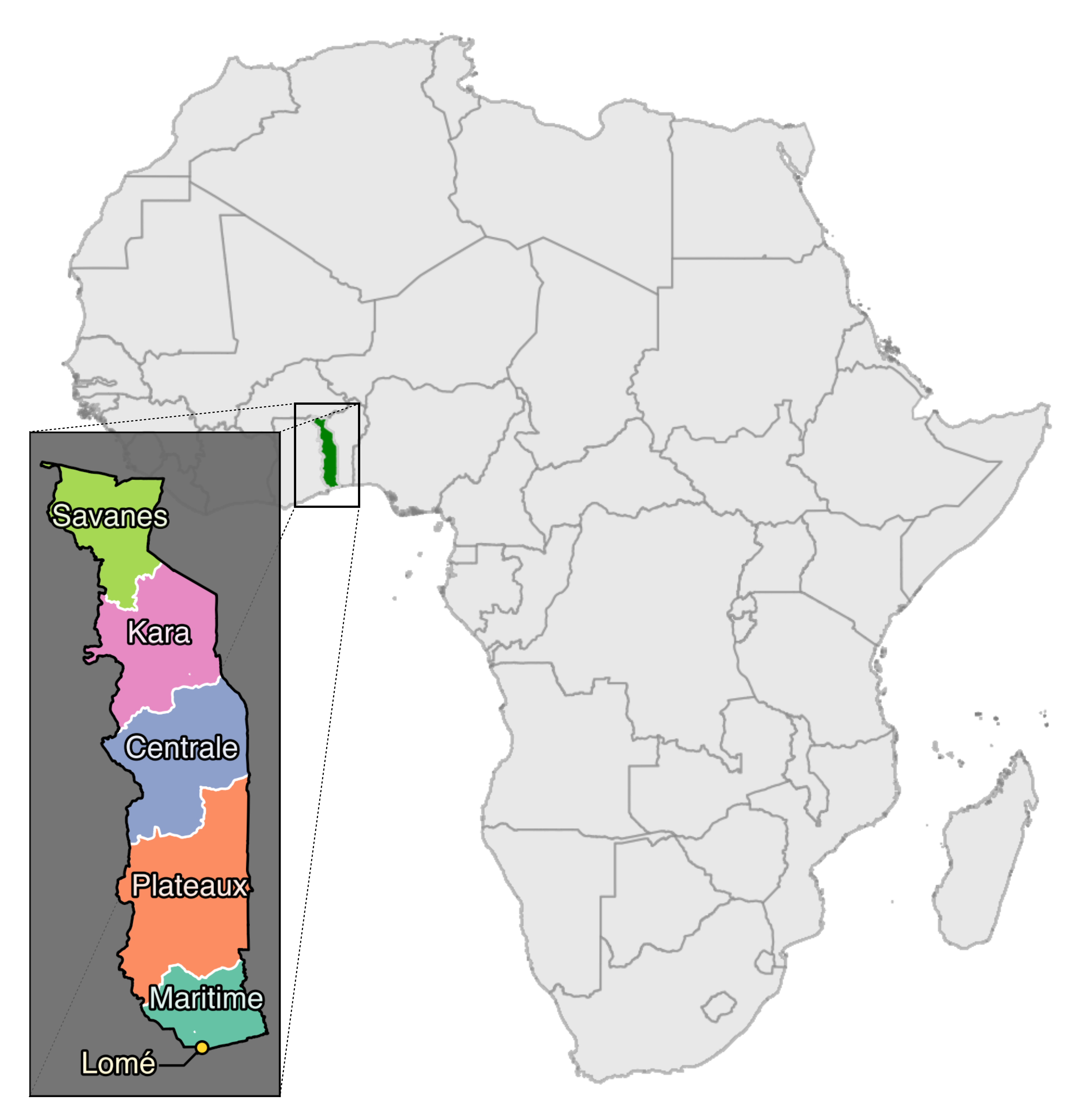}
\caption{Location of Togo (green) in Africa and major regions of the country (inset).}
\label{fig:togo_map}
\end{subfigure} \hfill
\begin{subfigure}[b]{0.595\textwidth}
\includegraphics[width=\textwidth,trim=0 0.3cm 0 0.39cm ,clip]{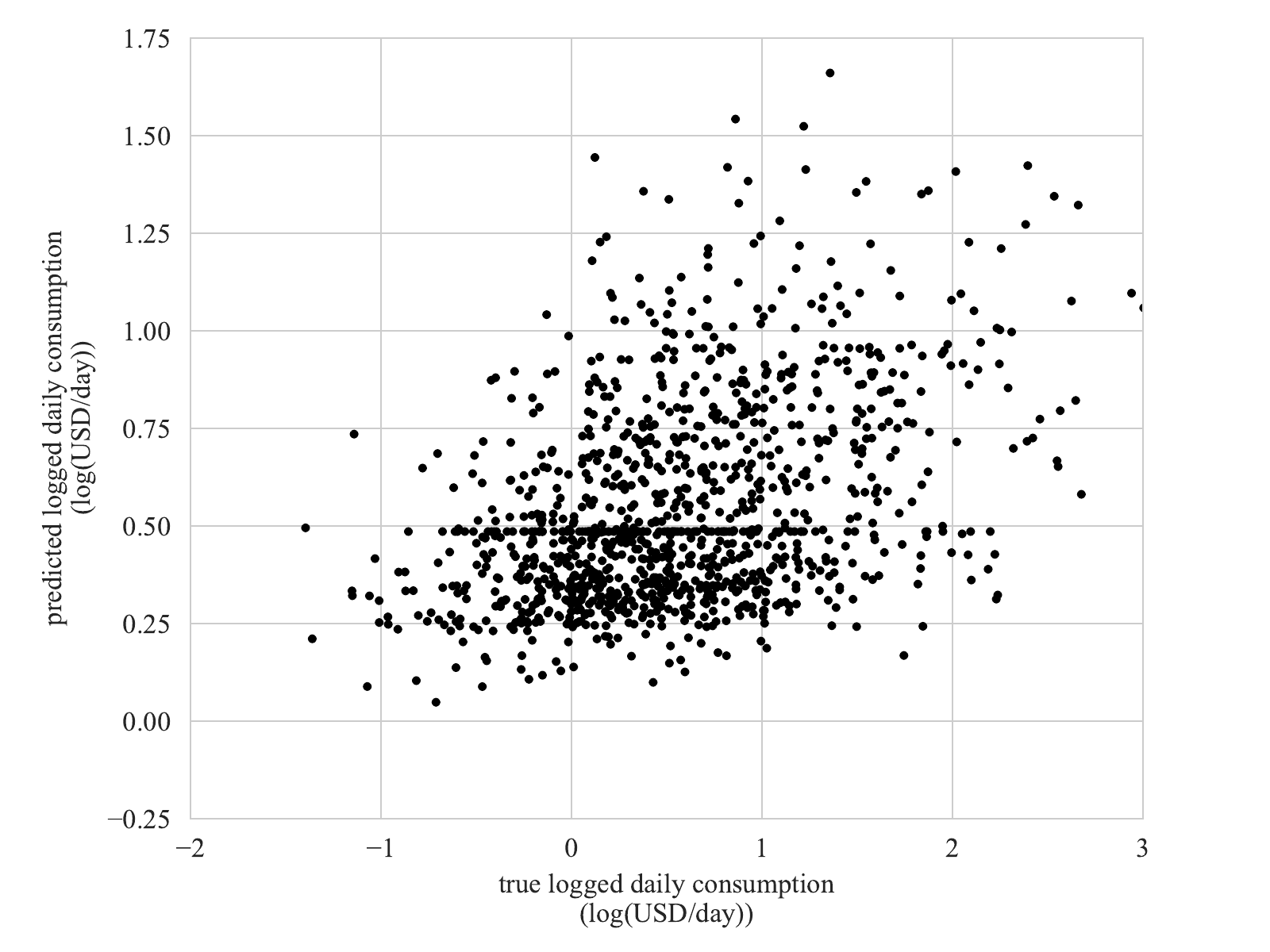}
\caption{True versus predicted log daily consumption for the holdout validation set, with a random forest trained on phone usage features.}
\label{fig:true_vs_predicted}
\end{subfigure}
\caption{Geographic context and poverty prediction model output for Togo.}
\label{fig:background}
\end{figure}

After an initial pilot of an emergency cash transfer program, named "Novissi", implemented through mobile money payments to residents in the greater metropolitan area surrounding the capital city of Lom\'e, the Togolese government sought to expand the program to 100 of Togo's poorest 397 cantons. Registrants were required to use a SIM card to register for the expanded program. Historical call records and cell phone usage metadata tied to those SIM cards were used to predict consumption, as illustrated in Figure \ref{fig:true_vs_predicted}.

\begin{figure}[ht]
\centering
\begin{subfigure}[b]{0.495\textwidth}
\includegraphics[width=\textwidth]{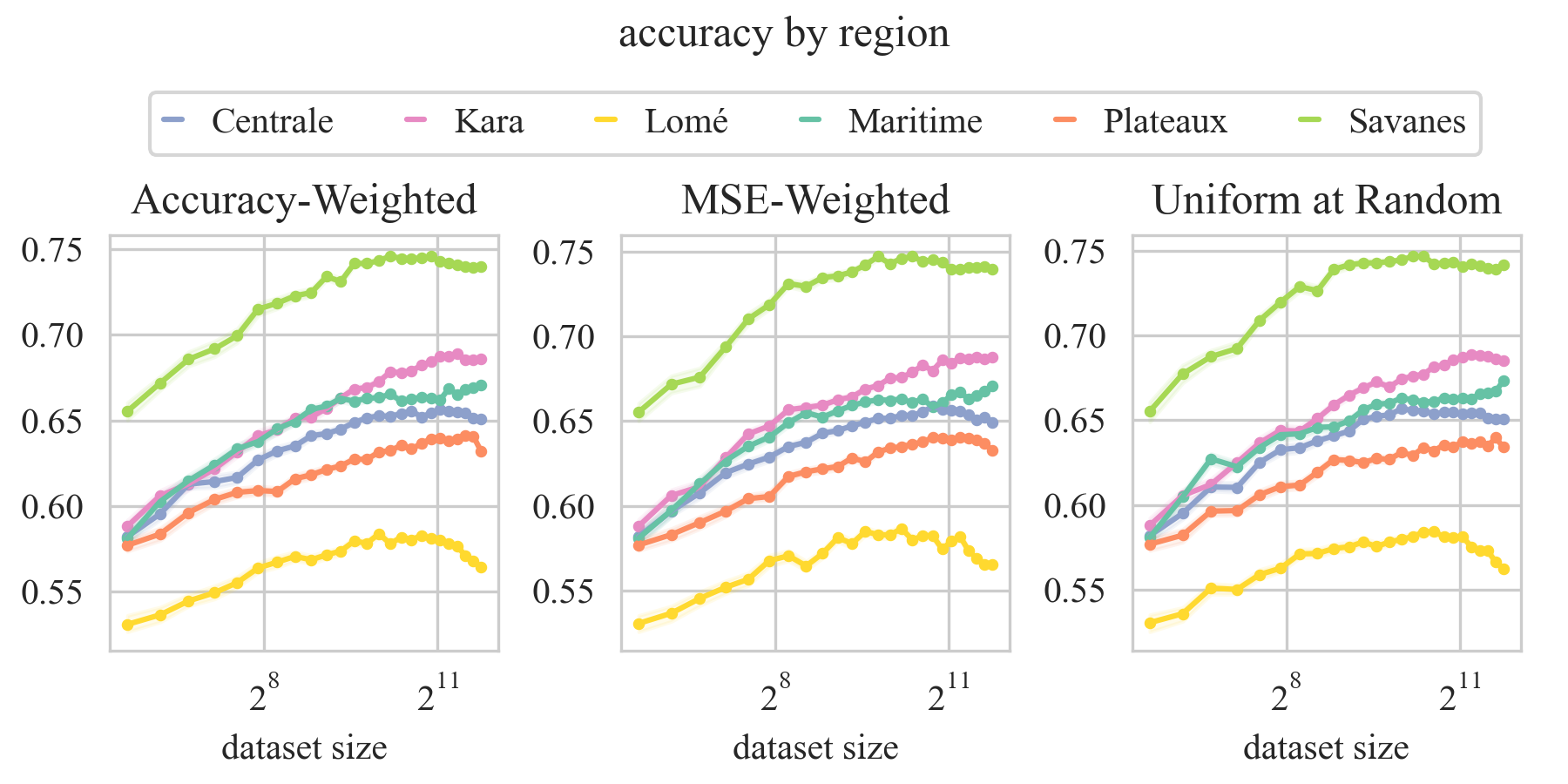}
\caption{Accuracy by region under sampling strategies.}
\label{fig:region_accuracy}
\end{subfigure} 
\begin{subfigure}[b]{0.495\textwidth}
\includegraphics[width=\textwidth]{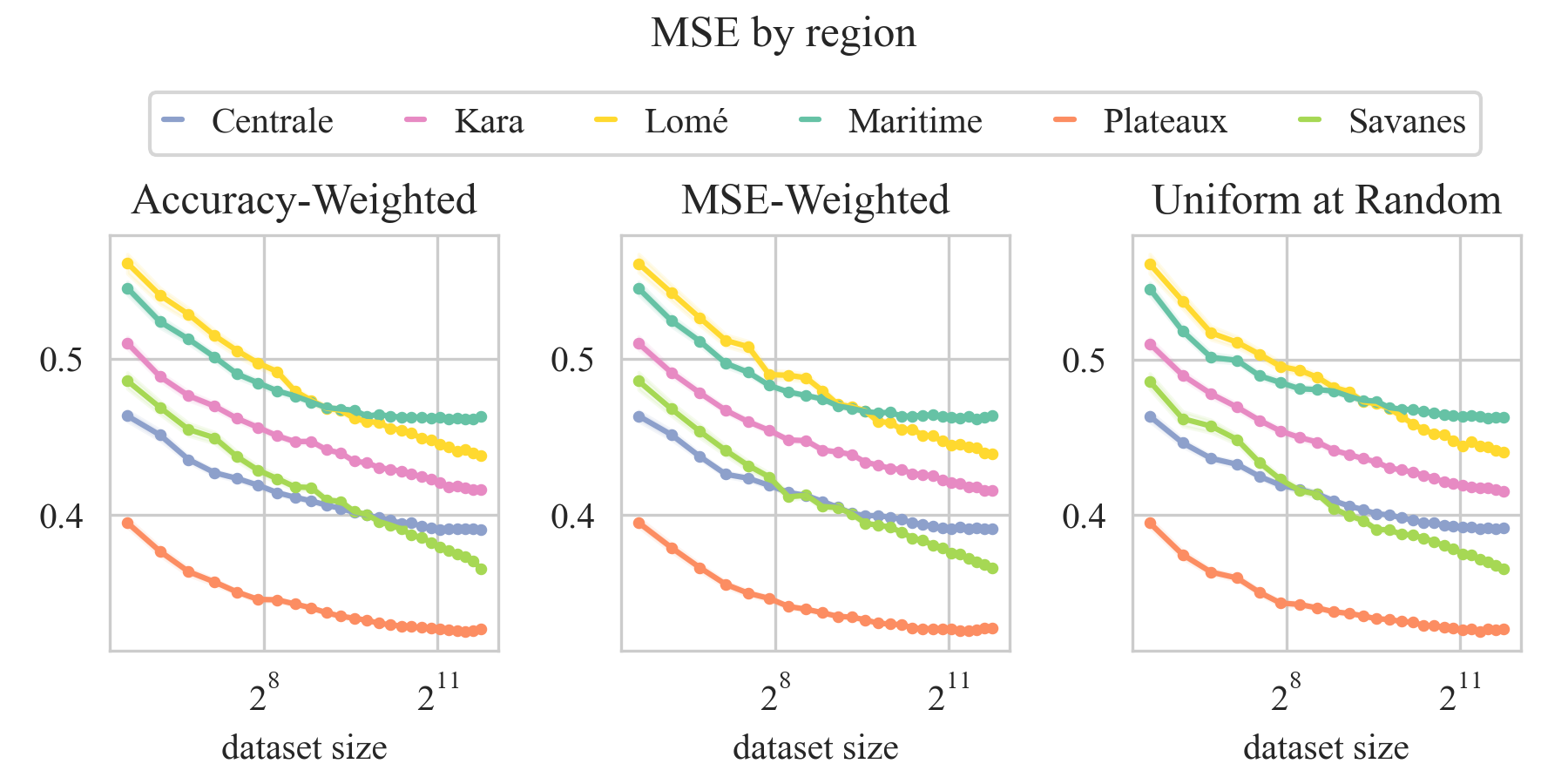}
\caption{MSE by region under sampling strategies.}
\label{fig:region_MSE}
\end{subfigure}
\caption{Per-group accuracy and MSE for each region of Togo. Note the differential performance of the model across regions, and the slight differences in the crossover points at which the model's performance for one subgroup improves compared to other subgroups, such as the MSE in Maritime vs Centrale.}
\label{fig:region_metrics}
\end{figure}

Given the geographic distribution of socioeconomic characteristics and potential differences in cell phone penetration and usage patterns across the country, it is not surprising that models trained using these data show differential performance across populations living in different regions (Fig.~\ref{fig:region_metrics}) (or other demographic breakdowns). \citet{aiken2022machine} found that aid targeting errors using this approach, despite the variance in performance, were lower than other, non-machine-learning-based approaches. Nevertheless, the difference in performances across regions was the starting point for our investigations into improving fairness characteristics, described below.

\section{Fairness Metrics Explained}\label{fairness_metrics_explained}
\subsection{Accuracy}
We follow the standard definition of accuracy, but calculate it for every group $g$ in the demographic category under consideration, $G$:

$$ A_g = \frac{\textrm{TP}_g + \textrm{TN}_g}{\textrm{TP}_g + \textrm{TN}_g + \textrm{FP}_g + \textrm{FN}_g} ~. $$

where TP$_g$, TN$_g$, FP$_g$, and FN$_g$ are the true positive, true negative, false positive, and false negative rates, respectively, calculated for group $g$.

The summary statistic, worst-case group accuracy, is the minimum over all groups: $\displaystyle \min_{g\in G} A_g$. Because we want to sample from the groups with the lowest accuracy, the weighting for each unseen data point $i$ in the label pool belonging to group $g$ is proportional to the group accuracy subtracted from one: $w_i \propto 1 - A_g$.

\subsection{Mean squared error}

As with accuracy, we calculate mean squared error for each demographic group:
$$ \textrm{MSE}_g = \frac{1}{|g|} \sum_{i \in g} \left(y_i - \hat{y}_i\right)^2 ~.$$

The summary statistic across groups, worst-case group MSE, is the maximum, and the sampling weight for each point $i$ is directly proportional to the group MSE. 

\subsection{Demographic disparity and absolute demographic deviance}
The final fairness metric we present is based on a measure of targeting exclusion developed in \citet{aiken2022machine}, namely group demographic parity (DP)\footnote{This definition of demographic parity differs from other definitions in the relevant literature.}: defined as the proportion of each group predicted to live below the US\$ 1.90 threshold, minus the true proportion of the group living below the same threshold:
$$ \textrm{DP}_g = \frac{\left(\textrm{TP}_g + \textrm{FP}_g\right) - \left(\textrm{TP}_g + \textrm{FN}_g\right)}{N_g} ~.$$

A positive value for DP indicates a given group is over-represented in the model-based targeting, while a negative value means the group is insufficiently targeted for aid receipt. For the disparity weighting strategy, the sampling weight is, like accuracy, proportional to one minus the group demographic parity value: $w_i \propto 1 - \textrm{DP}_g$.

For a summary statistic, we define absolute demographic deviance (ADD) as the sum of the absolute values of the demographic parity for each group, with the intent of capturing both errors of inclusion as well as errors of exclusion in a single number:
$$ \textrm{ADD} = \sum_{g\in G} \left|\textrm{DP}_g\right| ~.$$

\section{Dimensionality Reduction and Cross-Validation}
\label{app:additional_experiments}
\subsection{Dimensionality reduction}
\begin{figure}[ht]
\centering
\includegraphics[width=0.75\textwidth]{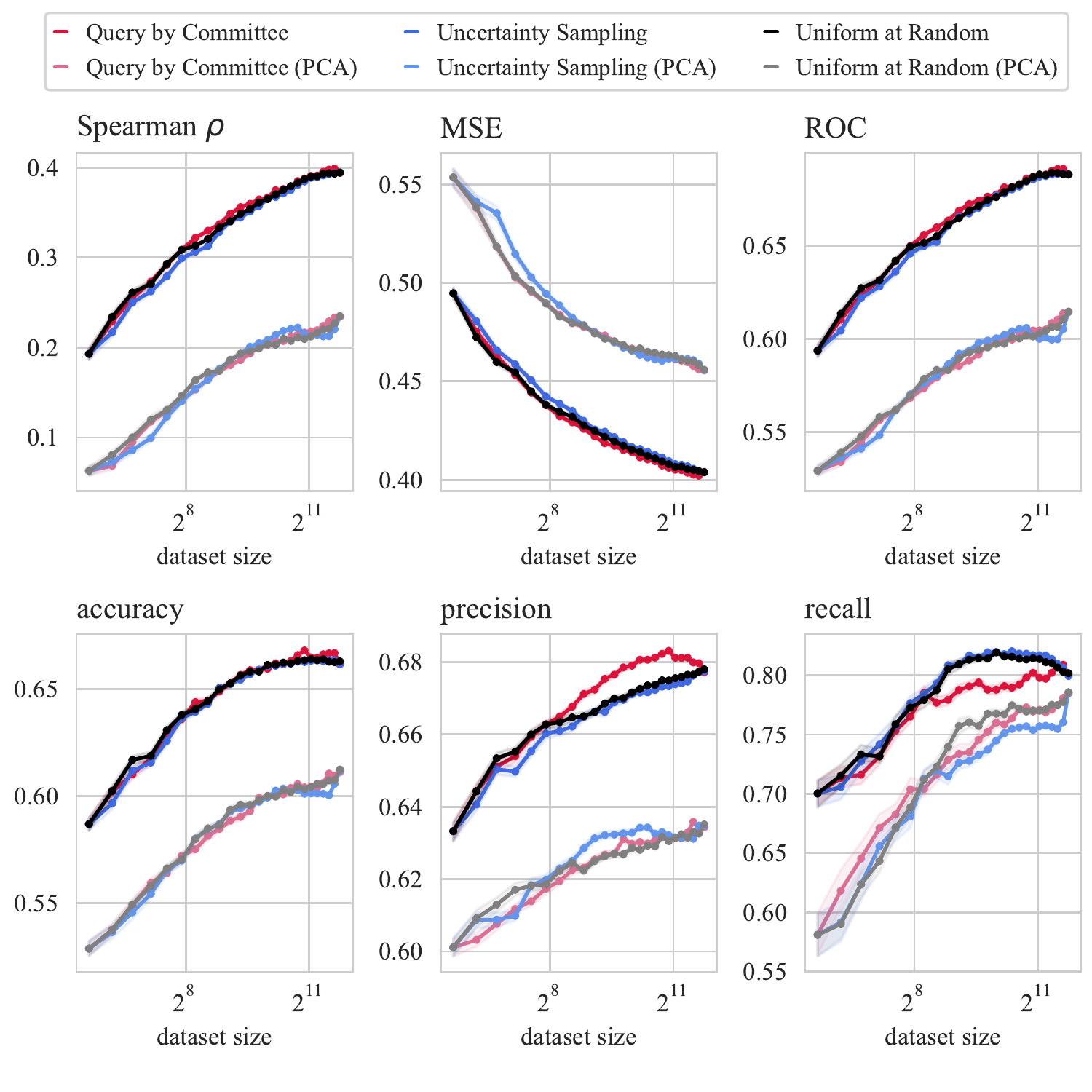}
\caption{Comparison of point-based strategies on the original data (darker colors) versus a PCA-compressed representation of the data (lighter colors). }
\label{fig:PCA_point_sampling_metrics}
\end{figure}

One potential issue with active learning approaches is that the dimension of the feature space makes it difficult to isolate the important dimensions along which new point selection would be beneficial \cite{tifreauniform}. To test this hypothesis in our context, we used a principal components analysis to reduce the dimension of our feature space from approximately 850 features to 50. As seen in Figure \ref{fig:PCA_point_sampling_metrics}, no appreciable gains were seen to using strategies other than uniform-at-random sampling in the PCA dataset (lighter colors), either.

\subsection{Cross-validation}
To maintain a simpler experimental setup, we skipped the usual cross-validation steps on models trained at each incremental dataset size $S_t$. To ensure that the lack of cross-validation did not artificially constrain the model performance, we present the results of cross-validating the model using three-fold cross-validation on the training set to find optimal tree depth at each incremental dataset size. As with dimensionality reduction, cross-validation did not show appreciable gains to model characteristics.
\begin{figure}[ht]
\centering
\includegraphics[width=0.75\textwidth]{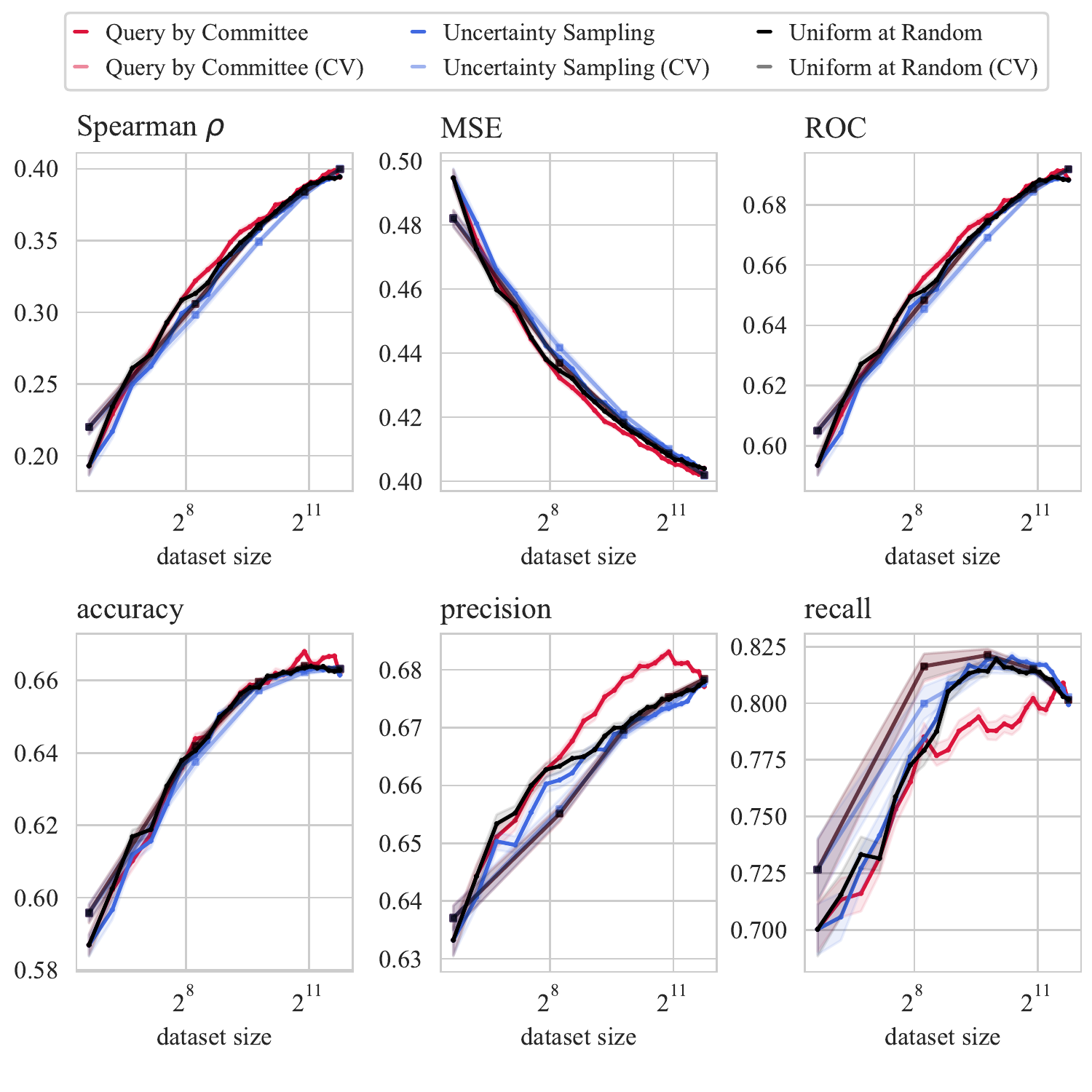}
\caption{Comparison of point-based strategies without cross-validation (darker colors, circular markers) versus the same strategies, with three-fold cross-validation run at each incremental dataset size (lighter colors, square markers).}
\label{fig:CV_point_sampling_metrics}
\end{figure}

\end{document}